\documentclass[10pt,twocolumn,letterpaper]{article}

\usepackage{cvpr}
\usepackage{times}
\usepackage{epsfig}
\usepackage{graphicx}
\usepackage{amsmath}
\usepackage{amssymb}
\usepackage{booktabs}
\usepackage{array}
\usepackage{multirow}
\usepackage{bm}
\usepackage[american]{babel}
\usepackage{microtype}
\usepackage{enumitem}
\usepackage{amsfonts}
\usepackage{mathrsfs}
\usepackage{epstopdf}
\usepackage{algorithm}
\usepackage{algorithmic}

\def\ie{{\em i.e.}}
\def\eg{{\em e.g.}}

\usepackage[pagebackref=true,breaklinks=true,letterpaper=true,colorlinks,bookmarks=false]{hyperref}

\cvprfinalcopy % *** Uncomment this line for the final submission

\def\cvprPaperID{5378} % *** Enter the CVPR Paper ID here

\ifcvprfinal\fi

\begin{document}

%%%%%%%%% TITLE
\title{Exploring Categorical Regularization for Domain Adaptive Object Detection}

\author{Chang-Dong Xu$^*$ \qquad Xing-Ran Zhao\thanks{Contributed equally.} \qquad Xin Jin \qquad Xiu-Shen Wei\thanks{X.-S. Wei is the corresponding author (weixs.gm@gmail.com). This research was supported by National Key R\&D Program of China (No. 2017YFA0700800).} \\
\noindent Megvii Research Nanjing, Megvii Technology\\
{\tt\small \{xuchangdong, zhaoxingran, jinxin\}@megvii.com, weixs.gm@gmail.com}\\
}

\maketitle
\thispagestyle{empty}

%%%%%%%%% ABSTRACT
\begin{abstract}
In this paper, we tackle the domain adaptive object detection problem, where the main challenge lies in significant domain gaps between source and target domains. Previous work seeks to plainly align image-level and instance-level shifts to eventually minimize the domain discrepancy. However, they still overlook to match crucial image regions and important instances across domains, which will strongly affect domain shift mitigation. In this work, we propose a simple but effective categorical regularization framework for alleviating this issue. It can be applied as a plug-and-play component on a series of Domain Adaptive Faster R-CNN methods which are prominent for dealing with domain adaptive detection. Specifically, by integrating an image-level multi-label classifier upon the detection backbone, we can obtain the sparse but crucial image regions corresponding to categorical information, thanks to the weakly localization ability of the classification manner. Meanwhile, at the instance level, we leverage the categorical consistency between image-level predictions (by the classifier) and instance-level predictions (by the detection head) as a regularization factor to automatically hunt for the hard aligned instances of target domains. Extensive experiments of various domain shift scenarios show that our method obtains a significant performance gain over original Domain Adaptive Faster R-CNN detectors. Furthermore, qualitative visualization and analyses can demonstrate the ability of our method for attending on the key regions/instances targeting on domain adaptation. Our code is open-source and available at \url{https://github.com/Megvii-Nanjing/CR-DA-DET}.
\end{abstract}

%%%%%%%%% BODY TEXT

\section{Introduction}

\begin{figure}[ht]
\centering
\includegraphics[width=0.45\textwidth]{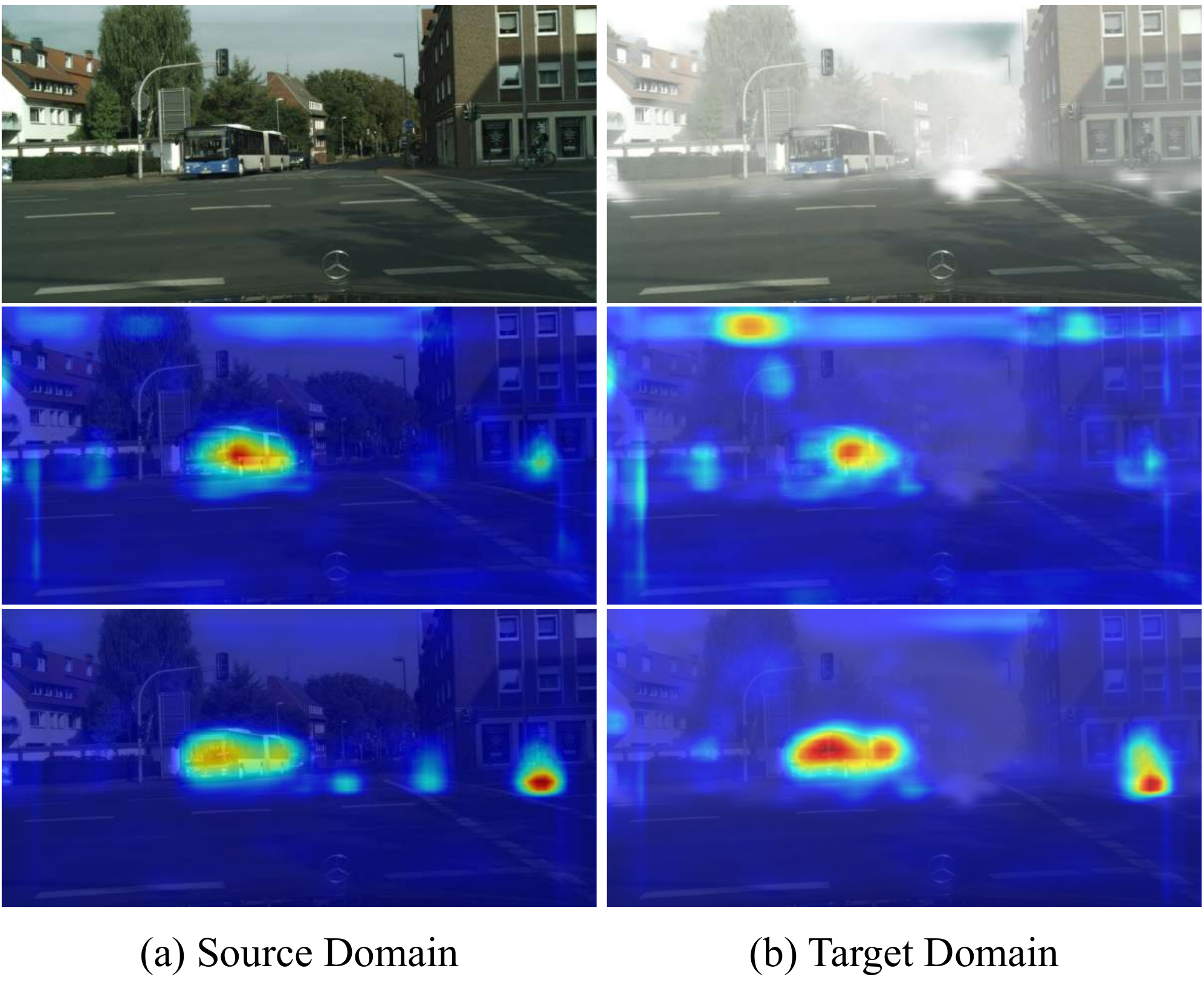}
\caption{\textbf{First row}: Exampled images from Cityscapes~\cite{cordts2016cityscapes} (source) and Foggy Cityscapes~\cite{sakaridis2018semantic} (target). \textbf{Second row}: Heatmaps by the backbone network (VGG-16~\cite{simonyan2015very}) of DA Faster R-CNN~\cite{chen2018domain}. \textbf{Third row}: Heatmaps by the backbone network of DA Faster R-CNN trained \emph{with} our categorical regularization framework. Our regularization framework enables more accurate alignment for crucial regions and important instances, and thus can assist the backbone network to activate the main objects of interest more accurately in \emph{both domains}, and lead to better adaptive detection performance.}
\label{fig:motivation}
%\vspace{-1em}
\end{figure}

Object detection is a fundamental task in computer vision, which aims to identify and localize objects of interest in an image. In the past decade, remarkable progress has been witnessed for object detection, with the advances of large-scale benchmarks~\cite{lin2014microsoft} and modern CNN-based detection frameworks, such as Fast/Faster R-CNN~\cite{girshick2015fast,ren2015faster}. However, state-of-the-art detectors require massive training images with bounding box annotations. This limits their generalization ability when facing new environments (\ie, the target domain) where the object appearance, background, and even weather condition significantly differ from the training images (\ie, the source domain). Meanwhile, due to the high cost of box annotations, it is not always feasible to acquire sufficient annotated training images from new environments.

\begin{figure*}[ht]
\centering
\includegraphics[width=0.9\textwidth]{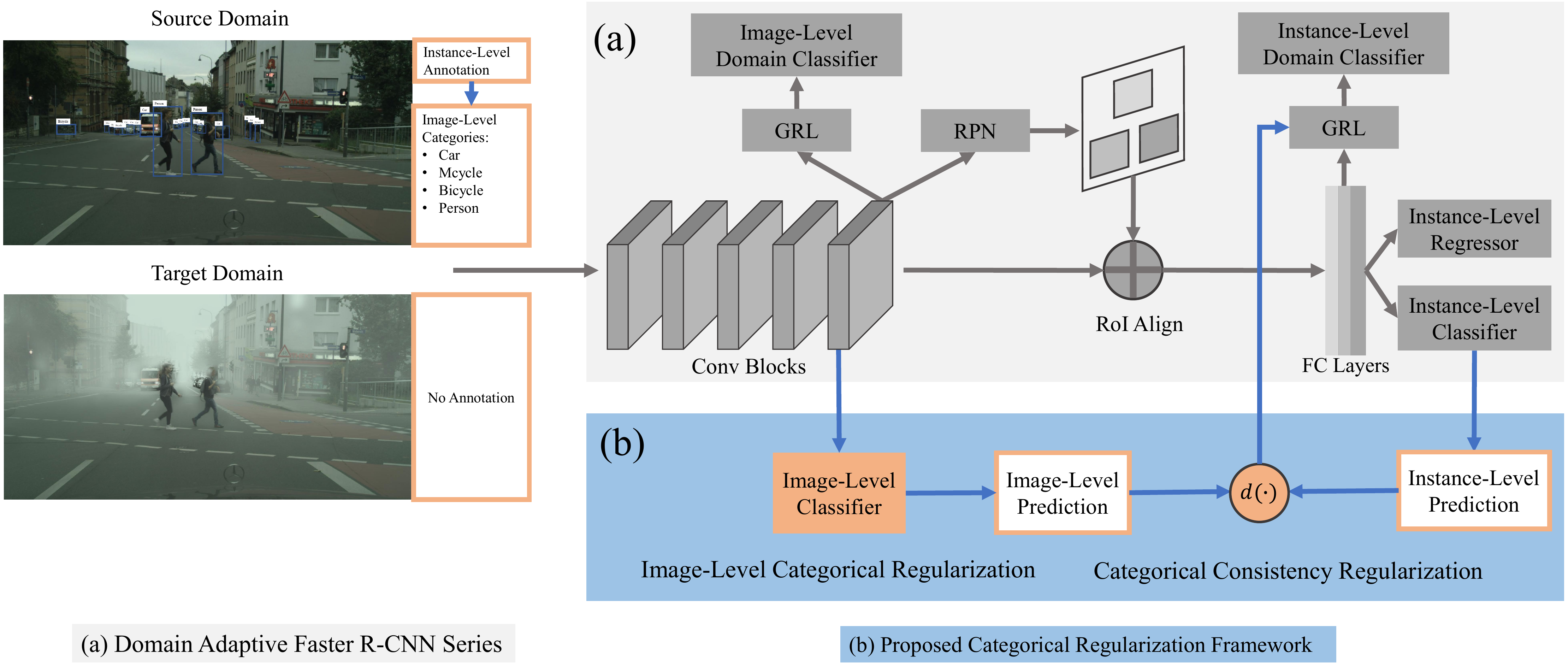}
\caption{Overview of our categorical regularization framework: a plug-and-play component for the Domain Adaptive Faster R-CNN series~\cite{chen2018domain,saito2019strong-weak}. Our framework consists of two modules, \ie, image-level categorical regularization (ICR) and categorical consistency regularization (CCR). The ICR module is an image-level multi-label classifier upon the detection backbone, which exploits the weakly localization ability of classification CNNs to obtain crucial regions corresponding to categorical information. The CCR module considers the consistency between the image-level and instance-level predictions as a novel regularization factor, which can be used to automatically hunt for hard aligned instances in the target domain during instance-level alignment.}
\label{fig:overview}
%\vspace{-1em}
\end{figure*}

In such situations, unsupervised domain adaptation offers an appealing solution by adapting object detectors from label-rich source domains to unlabeled target domains. Among a large number of methods, a promising manner for domain adaptation is to utilize the domain classifier to measure domain discrepancy, and train the domain classifier and feature extractor in an adversarial way~\cite{ganin2015unsupervised,tzeng2017adversarial}. In the literature, adversarial training has been well-studied for domain adaptive image classification~\cite{ganin2015unsupervised,ganin2016domain,long2015learning,tzeng2017adversarial}, semantic segmentation~\cite{hoffman2016fcns,sankaranarayanan2018learning,tsai2018learning} and object detection~\cite{chen2018domain,saito2019strong-weak,zhu2019adapting,he2019multi-adversarial}.

Among many domain adaptive detection methods, Domain Adaptive (DA) Faster R-CNN~\cite{chen2018domain} is the most representative work that integrates Faster R-CNN~\cite{ren2015faster} with adversarial training. To address the domain shift problem, it aligns both the image and instance distributions across domains with adversarial training. Recently, DA Faster R-CNN has rapidly evolved into a successful series~\cite{saito2019strong-weak,zhu2019adapting,he2019multi-adversarial,hsu2019progressive}. Specifically, Saito~{\em et~al.}~\cite{saito2019strong-weak} and Zhu~{\em et~al.}~\cite{zhu2019adapting} improved DA Faster R-CNN based on the observation that the plain image-level alignment forces to align non-transferable backgrounds, while the object detection task by nature focuses on local regions that may contain objects of interest. Furthermore, although instance-level alignment can match object proposals in both domains, current practices~\cite{chen2018domain,he2019multi-adversarial} lack the ability of identifying the hard aligned instances from excessive low-value region proposals.

Aiming at these issues, we propose a novel categorical regularization framework, which can assist the Domain Adaptive Faster R-CNN series~\cite{chen2018domain,saito2019strong-weak} to focus on aligning the crucial regions and important instances cross domains. Thanks to the accurate alignment for such regions and instances, the detection backbone networks can activate objects of interest more accurately in \emph{both domains} (cf. Figure~\ref{fig:motivation}), and thus lead to better adaptive object detection results.

Concretely, our framework consists of two regularization modules, \ie, image-level categorical regularization (ICR), and categorical consistency regularization (CCR) (cf. Figure~\ref{fig:overview}). For image-level categorical regularization, we attach the detection backbone network with an image-level multi-label classifier, and train it with categorical supervisions from the source domain. The classification manner enables the backbone to learn object-level concepts from the holistic images, without being affected by the distribution of non-transferable source backgrounds~\cite{zhou2015object,zhou2016learning}. It allows us to implicitly align the crucial regions on both domains at the image level. For categorical consistency regularization, we take into account the consistency between image-level predictions by the attached classifier and instance-level predictions by the detector. We adopt this categorical consistency as a novel regularization factor, and use it to increase the weights of the hard aligned instances in the target domain during instance-level alignment.

The main contributions of this work are three-fold:
\begin{itemize}
	\item We present a novel categorical regularization framework for domain adaptive object detection, which can be applied as a plug-and-play component for the prominent Domain Adaptive Faster R-CNN series. Our framework is cost-free as requiring no further annotations, and also hyperparameter-free for performing on the vanilla detectors.
	\item We design two regularization modules, by exploiting the weakly localization ability of classification CNNs and the categorical consistency between image-level and instance-level predictions. They enable us to focus on aligning object-related regions and hard aligned instances that are directly pertinent to object detection.
	\item We conduct extensive experiments of various domain shift scenarios to validate the effectiveness of our categorical regularization framework. Our framework can significantly boost the performance of existing Domain Adaptive Faster R-CNN detectors~\cite{chen2018domain,saito2019strong-weak}, and produce state-of-the-art results on benchmark datasets.
\end{itemize}

\section{Preliminaries and Related Work}

\subsection{CNN-based Object Detection}  In the past few years, the rise of deep convolutional neural networks led to a sharp paradigm shift of object detection~\cite{liu2019deep}. Among a large number of approaches, the two-stage R-CNN series~\cite{girshick2014rich,girshick2015fast,ren2015faster,lin2017feature} have become the mainstream detection framework. The pioneer work, \ie, R-CNN~\cite{girshick2014rich}, extracts region proposals from the image with low-level vision techniques~\cite{uijlings2013selective}, and applies a network to classify each region of interest (RoI) independently. Fast R-CNN~\cite{girshick2015fast} improves R-CNN by sharing convolutional features among RoIs, and thus enables fast training and inference. Faster R-CNN~\cite{ren2015faster} advances the region proposal generation process with a Region Proposal Network (RPN). RPN shares the feature extraction backbone with the detection head, which in essence is a Fast R-CNN~\cite{girshick2015fast}. Faster R-CNN is a famous two-stage detection framework, and is the foundation for many follow-up works~\cite{gidaris2015object,dai2016r,lin2017feature}. While recently single-stage detectors have emerged as a popular paradigm~\cite{redmon2016you,liu2016ssd,lin2017focal}, many top-performing systems still adopt the proven two-stage pipeline~\cite{lin2017feature,he2017mask}. 

Thanks to the flexibility of Faster R-CNN, recently, it is widely adapted for domain adaptive object detection~\cite{chen2018domain,saito2019strong-weak,zhu2019adapting,he2019multi-adversarial} with adversarial training~\cite{ganin2015unsupervised}. Other approaches, such as self-training~\cite{kim2019self-training,roychowdhury2019automatic}, are also exploited for domain adaptive object detection in the literature.

\subsection{Domain Adaptive Faster R-CNN Series}\label{subsec:DA_Faster}
Domain Adaptive (DA) Faster R-CNN~\cite{chen2018domain} is a prominent two-stage object detector for dealing with the challenging domain adaptive object detection problem.
It is an intuitive extension of Faster R-CNN~\cite{ren2015faster}, which aligns both the image and instance distributions by learning domain classifiers in an adversarial manner. For the image-level alignment, the domain classifier is trained on each activation (channel-wise descriptor) from the feature map after the base convolutional layers, while for instance-level alignment, the domain classifier is trained with instance-level RoI features. Furthermore, the consistency between image-level and instance-level domain classifiers is enforced to learn the cross-domain robustness for RPN.

Formally, for a given image, let $D\!=\!0$ denote that it is from the source domain while $D\!=\!1$ denote that it is from the target domain. Let $\hat{D}^{(u,v)}$ denote the output of the image-level domain classifier for the activation located at $(u,v)$ of the feature map, then the image-level alignment loss can be written as
\begin{equation}\label{eq:loss_im_ad}
 \mathcal{L}_{img} \!=\! -\! \sum_{u,v}\! \left[D\! \log \! \hat{D}^{(u,v)} \!+\! (1\!-\!D)\! \log (1\!-\!\hat{D}^{(u,v)}) \! \right].
\end{equation}
Let $\hat{D}_{j}$ denote the output of the instance-level domain classifier for the $j$-th region proposal, then the instance-level alignment loss is as follows
\begin{equation}\label{eq:loss_in_ad}
 \mathcal{L}_{ins} = - \sum_{j} \left[D \log \hat{D}_{j} + (1-D) \log (1-\hat{D}_{j}) \right].
\end{equation}
Furthermore, let $\mathcal{L}_{cst}$ denote the consistency loss for image-level and instance-level domain classifiers, and let $\mathcal{L}_{det}$ be the original training loss for Faster R-CNN~\cite{ren2015faster}. The overall objective $\mathcal{L}_{DAF}$ for DA Faster R-CNN can be written as
\begin{equation}\label{eq:overall_loss_DAF}
 \mathcal{L}_{DAF} = \mathcal{L}_{det} + \lambda \cdot (\mathcal{L}_{img} + \mathcal{L}_{ins} + \mathcal{L}_{cst}),
\end{equation}
where $\lambda$ is a hyper-parameter to balance the detection loss and the domain adaptation components. The adversarial training for adaptation components is implemented by the gradient reverse layer (GRL)~\cite{ganin2015unsupervised}, where the sign of gradients is flipped when training the base convolutional layers.

As aforementioned, DA Faster R-CNN~\cite{chen2018domain} may fail to align the crucial regions and important instances which are crucial for adaptive detection. Meanwhile, it tends to fit the distribution of non-transferable source backgrounds, as the training process
involves a large amount of background proposals. Recent works attempted to improve DA Faster R-CNN by replacing the plain image-level alignment model with a weak alignment model~\cite{saito2019strong-weak} or a region-level alignment model~\cite{zhu2019adapting}, and found that the instance-level alignment model is not necessary in presence of other local alignment model~\cite{saito2019strong-weak}. We term to these methods collectively as Domain Adaptive Faster R-CNN series.

A high-level diagram of Domain Adaptive Faster R-CNN series is shown in Figure~\ref{fig:overview} (a), where we follow the paradigm of DA Faster R-CNN~\cite{chen2018domain} but omit the part of $\mathcal{L}_{cst}$ which is not an essential ingredient in our regularization framework. Please note that Figure~\ref{fig:overview} (a) is a conceptual diagram, and not all components of the Domain Adaptive Faster R-CNN series strictly follow this structure.

\subsection{Weakly Localization by Classification CNNs}
\begin{figure}[t]
\centering
\includegraphics[width=0.45\textwidth]{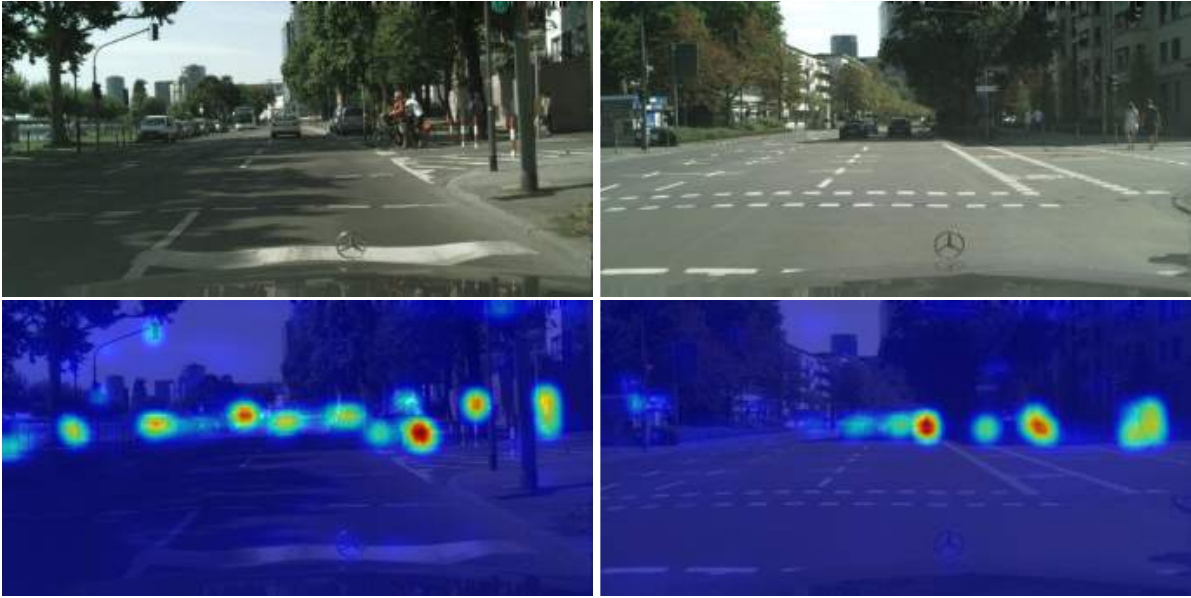}
\caption{Visualization of the weakly localization ability of multi-label classification CNNs. The CNN model is VGG-16 trained on Cityscapes~\cite{cordts2016cityscapes}.}
\label{fig:cnn_attention}
\end{figure}

It is widely acknowledged that CNNs trained for single-label image classification tend to produce high responses on the local regions containing the main objects~\cite{zeiler2014visualizing,zhou2016learning,zhou2015object}. Analogously, CNNs trained for multi-label classification also have the weakly localization ability for the objects associated with image-level categories~\cite{wang2016cnn,wei2015hcp}.

Taking the Cityscapes~\cite{cordts2016cityscapes} dataset for an example, we collect all instance-level labels into an image-level label vector, and train VGG-16~\cite{simonyan2015very} for multi-label image classification. Figure~\ref{fig:cnn_attention} shows the heatmaps for two exampled images from Cityscapes, where the main objects related to image-level categories such as ``car'', ``person'' and ``rider'' are weakly localized.

\section{Approach}
%In this section, we elaborate our categorical regularization framework for Domain Adaptive Faster R-CNN series.

\subsection{Framework Overview}

The overview of our categorical regularization framework is illustrated in Figure~\ref{fig:overview}. In general, our framework improves the DA Faster R-CNN series detectors~\cite{chen2018domain,saito2019strong-weak,he2019multi-adversarial} by exploring categorical regularization from two aspects: image-level categorical regularization (ICR) and categorical consistency regularization (CCR). Note that the ICR module does not depend on the CCR module, and thus it can be individually integrated with DA Faster R-CNN detectors which only perform image-level alignment~\cite{saito2019strong-weak}. 

Our framework enables better alignment of crucial regions and important instances across domains. Consequently, the detection backbone produces more accurate activations on objects of interest of both domains (cf. Figure~\ref{fig:motivation}), leading to better adaptive detection performance. Our framework is flexible and generalizable -- it does not depend on specific algorithms for either image or instance alignment.

\subsection{Image-Level Categorical Regularization}

Image-level categorical regularization (ICR) is exploited to obtain the sparse but crucial image regions corresponding to categorical information. We achieve this with a weakly supervised solution, which can learn discriminative features for objects of interest, without being affected by the distribution of non-transferable source backgrounds. While the standard training for Faster R-CNN can learn discriminative features for objects of interest, it tends to fit the source backgrounds due to the large amount of background RoIs sampled for training. Since the patterns of source backgrounds are non-transferable, plain image-level alignment may lead to noisy activations in target domains (cf. Figure~\ref{fig:motivation}).

In our proposal, as illustrated in Figure~\ref{fig:overview} (b), we attach the detection backbone with an image-level multi-label classifier, and train it with supervisions from the source domain. Such categorical supervisions are cost-free for detection datasets, and can be easily acquired by collecting all instance-level categories in an image into an image-level categorical vector.

Given the detection backbone network, we perform global average pooling on the output of the last convolutional layer, and feed the pooled features into a plain multi-label classifier implemented by a 1$\times$1 convolution. We train this image-level classifier with the standard cross-entropy multi-label loss by
\begin{equation}\label{eq:loss_ICR}
 \mathcal{L}_{ICR}= \sum\nolimits_{c=1}^{C}y^{c}\log(\hat{y}^{c}) + (1-y^{c})\log(1-\hat{y}^{c}),
\end{equation}
where $C$ is the total number of categories of a detection dataset, $y^{c}$ is the ground truth label, and $\hat{y}^{c}$ is the predicted one. $y^{c}=1$ denotes that there is at least one object of category $c$ appearing in this image, while $y^{c}=0$ means there is no object of category $c$ in the image.

The image-level categorical supervisions encourage the detection backbone to learn category-specific features that can activate object-related regions. This allows us to align the crucial regions of both domains with an image-level alignment model (\eg, Equation~\eqref{eq:loss_in_ad}). Meanwhile, because there is no background supervision involved in the training process of our image-level multi-label classifier, the risk of fitting (even over-fitting) non-transferable source backgrounds is greatly reduced.

\subsection{Categorical Consistency Regularization}
We design a categorical consistency regularization (CCR) module to automatically  hunt for the hard aligned instances in target domains. Our motivation lies in two aspects. First, current instance alignment models~\cite{chen2018domain,he2019multi-adversarial} may be dominated by the excessive low-value background proposals, as they can not identify the hard foreground instances in the target domain. Second, the attached image-level classifier and the instance-level detection head are complementary, because the former exploits the whole image-level context while the latter enjoys more accurate RoI features.

Building upon those above considerations, we adopt the categorical consistency between the image-level and instance-level predictions as a measure for the hardness of classifying a certain target instance. Intuitively, if the image-level classifier predicts that there is no ``person'' in a target image while the detection head classifies a certain instance as ``person'', this instance should be a hard but informative sample for current detection model. Therefore, we utilize this consistency as a regularization factor to increase the weight of hard aligned samples in target domains during instance-level alignment.

Specifically, assume that the detection head classifies the $j$-th instance in a target image as category $c$, we let $\hat{p}^{c}_{j}$ denote the estimated probability. Using the notation in Equation~\eqref{eq:loss_ICR}, we let $\hat{y}^{c}$ denote the image-level estimation of the probability that this image contains objects of category $c$. We define the following distance function to measure the categorical consistency between the instance-level and image-level predictions as
\begin{equation}\label{eq:CCR}
 \mathit{d}_{j} = e^{|\hat{p}^{c}_{j} - \hat{y}^{c}|}.
\end{equation}
Here the exponent form characterizes the intuition that while a small disagreement may come from the model's variance, a large disagreement should be attributed to the hardness in classifying this instance.

We use Equation~\eqref{eq:CCR} to weight the instance-level adversarial loss, which in implementation is equivalent to weight the gradients passed through the gradient reversal layer (GRL) during training. Take the instance alignment model (\ie, Equation~\eqref{eq:loss_in_ad}) in DA Faster R-CNN~\cite{chen2018domain} for an example, the instance-level alignment loss with CCR can be written as
\begin{equation}\label{eq:loss_in_ad_CCR}
 \mathcal{L}_{ins}^{CCR} \!=\! - \sum_{j}  \mathit{d}_{j} \left[D \log \hat{D}_{j} \!+\! (1\!-\!D) \log (1\!-\!\hat{D}_{j}) \right].
\end{equation}

It is worth noting that, we only apply Equation~\eqref{eq:CCR} to weight \emph{foreground} instances from the target domain, according to the predictions of detection head. We keep the weights for source instances and the background instances from the target domain unchanged (\ie, $d_j=1$), as the former have supervision signals from the source domain, while the latter are not as important as foreground proposals.

\subsection{Integration with DA Faster R-CNN Series}
In this work, we take the DA Faster R-CNN~\cite{chen2018domain} and the state-of-the-art strong-weak aligned Faster R-CNN~\cite{saito2019strong-weak} as our baseline detectors. In the following, we term them as ``DA-Faster'' and ``SW-Faster'' for simplicity. In fact, other Domain Adaptive Faster R-CNN detectors~\cite{he2019multi-adversarial,zhu2019adapting} may also be compatible with our framework with minor modifications.

\paragraph{Integration with DA-Faster.} Integrating our framework with DA-Faster~\cite{chen2018domain} is straightforward. We attach an image-level multi-label classifier to the backbone, by adding a global averaging pooling layer and a 1$\times$1 convolution layer. Furthermore, we use our CCR to weight the gradients passed through the reverse gradient layer (GRL) for instance-level alignment. The modified overall objective of DA-Faster with our regularization framework can be written as
\begin{equation}\label{eq:overall_loss_DAF_ours}
 \mathcal{L}_{DAF}^{*} \!=\! \mathcal{L}_{det} \!+\! {\mathcal{L}_{ICR}} \!+ \lambda \cdot (\! \mathcal{L}_{img}  \!+ \! \mathcal{L}_{ins}^{CCR} \!+\! \mathcal{L}_{cst}),
\end{equation}
where $\lambda$ is set to $0.1$ in~\cite{chen2018domain}, and our method does not introduce additional hyper parameters.

\paragraph{Integration with SW-Faster.} SW-Faster~\cite{saito2019strong-weak} improves the strong image-level alignment model of DA-Faster with a weak global alignment model, and replaces the instance-level alignment model with a strong local alignment model. Since our categorical regularization framework is independent of the specific algorithms for alignment, our ICR module can be straightly integrated into SW-Faster. Furthermore, we add an instance-level alignment model, which is the same to that of DA-Faster, into the pipeline of SW-Faster during training. This allows us to apply our CCR module to further improve SW-Faster. The modified overall objective for SW-Faster with our regularization framework can be written as
\begin{equation}\label{eq:overall_loss_SWF_ours}
 \mathcal{L}_{SWF}^{*} \!=\! \mathcal{L}_{det} \!+\! \mathcal{L}_{ICR} \!+ \lambda' \cdot (\! \mathcal{L}_{ins}^{CCR} \!+\! \mathcal{L}_{global} \!+\! \mathcal{L}_{local} ),
\end{equation}
where $\lambda'$ is set to $1.0$, and $\mathcal{L}_{global}$ and $\mathcal{L}_{local}$ denote the global alignment loss and local alignment loss in~\cite{saito2019strong-weak}.

\section{Experiments}

%In this section, we conduct extensive experiments of various domain shift scenarios to validate the effectiveness of our categorical regularization framework. We also present visualization and analyses to verify the ability of our method for aligning the key regions/instances of both domains.

\subsection{Empirical Setup}
\begin{table*}[!ht]
\centering
\footnotesize
\caption{\textbf{Weather Adaptation}: Results on Foggy Cityscapes, using models trained on Cityscapes.}
\begin{tabular}{l|p{1cm}<{\centering}p{1cm}<{\centering}p{1cm}<{\centering}p{1cm}<{\centering}p{1cm}<{\centering}
p{1cm}<{\centering}p{1cm}<{\centering}p{1cm}<{\centering}|c}
\toprule
Method & person & rider & car & truck & bus & train & mcycle & bicycle & mAP \\
\hline
Faster R-CNN (Source) &  24.4 & 30.5  & 32.6 & 10.8  & 25.4 & 9.1 & 15.2 & 28.3 & 22.0 \\
MA-Faster~\cite{he2019multi-adversarial} &  28.4 & 39.5  & 43.9 & 23.8  & 39.9 & 33.3 & 29.2 & 33.9 & 34.0 \\
Selective-Faster~\cite{zhu2019adapting} &  33.5 & 38.0  & 48.5 & 26.5  & 39.0 & 23.3 & 28.0 & 33.6 & 33.8 \\
\hline
DA-Faster~\cite{chen2018domain} & 28.7 & 36.5  & 43.5 & 19.5  & 33.1 & 12.6 & 24.8 & 29.1 & 28.5 \\
\textbf{DA-Faster-ICR (Ours)} & 28.7 & 37.3  & 43.0 & 21.9  & 36.9 & 9.2 & 25.9 & 31.9 & 29.4 \\
\textbf{DA-Faster-ICR-CCR (Ours)} & 29.7 & 37.3 & 43.6  & 20.8 & 37.3  & 12.8 & 25.7 & 31.7 & 29.9 \\
SW-Faster~\cite{saito2019strong-weak} & 32.3 & 42.2  & 47.3 & 23.7  & 41.3 & 27.8 & 28.3 & 35.4 & 34.8 \\
\textbf{SW-Faster-ICR (Ours)} & \textbf{33.1} & \textbf{44.2}  & 48.8 & \textbf{27.7}  & 44.9 & 27.9 & 29.4 & \textbf{36.2} & 36.5 \\
\textbf{SW-Faster-ICR-CCR (Ours)} & 32.9 & 43.8  & \textbf{49.2} & 27.2  & \textbf{45.1} & \textbf{36.4} & \textbf{30.3} & 34.6 & \textbf{37.4} \\
\hline
Faster R-CNN (Oracle) & 36.2 & 47.7  & 53.0 & 34.7  & 51.9 & 41.0 & 36.8 & 37.8 & 42.4 \\
\bottomrule
\end{tabular}
\label{tbl:weather_ad}
\end{table*}

\begin{table*}[!ht]
\centering
\footnotesize
\caption{\textbf{Scene Adaptation}: Results of \textbf{7 common categories} on the \emph{daytime} subset of BDD100k, using models trained on Cityscapes.}
\begin{tabular}{l|p{1cm}<{\centering}p{1cm}<{\centering}p{1cm}<{\centering}p{1cm}<{\centering}p{1cm}<{\centering}p{1cm}<{\centering}p{1cm}<{\centering}p{1cm}<{\centering}|c} 
\toprule
Method & person & rider & car & truck & bus & train & mcycle & bicycle & mAP \\
\hline
Faster R-CNN (Source) &  26.9 & 22.1 & 44.7 & 17.4  & 16.7 & - & 17.1 & 18.8 & 23.4 \\
\hline
DA-Faster~\cite{chen2018domain} & 29.4 & 26.5  & 44.6 & 14.3  & 16.8 & - & 15.8 & 20.6 & 24.0 \\
\textbf{DA-Faster-ICR (Ours)} & 29.1 & 28.6  & 44.8 & 14.9  & 15.8 & - & 17.1 & 22.4 & 24.7 \\
\textbf{DA-Faster-ICR-CCR (Ours)} & 29.3 & 28.4  & 45.3 & 17.5  & 17.1 & - & 16.8 & 22.7 & 25.3 \\
SW-Faster~\cite{saito2019strong-weak} & 30.2 & 29.5  & 45.7 & 15.2  & 18.4 & - & 17.1 & 21.2 & 25.3 \\
\textbf{SW-Faster-ICR (Ours)} & 30.9 & 31.2 & 45.6 & 15.9  & 18.4 & - & \textbf{19.3} & 23.7 & 26.4 \\
\textbf{SW-Faster-ICR-CCR (Ours)} & \textbf{31.4} & \textbf{31.3}  & \textbf{46.3} & \textbf{19.5}  & \textbf{18.9} & - & 17.3 & \textbf{23.8} & \textbf{26.9} \\
\hline
Faster R-CNN (Oracle) & 35.3 & 33.2  & 53.9 & 46.3  & 46.7 & - & 25.6 & 29.3 & 38.6 \\
\bottomrule
\end{tabular}
\label{tbl:scene_ad}
\end{table*}

\paragraph{Datasets.}
Five public datasets are utilized in our experiments, including Cityscapes~\cite{cordts2016cityscapes}, Foggy Cityscapes~\cite{sakaridis2018semantic}, BDD100k~\cite{yu2018bdd100k}, PASCAL VOC~\cite{everingham2010pascal}, and Clipart1k~\cite{inoue2018cross-domain}.
\begin{itemize}
\setlength{\itemsep}{1pt}
\setlength{\parsep}{0pt}
\setlength{\parskip}{0pt}

	\item \textbf{Cityscapes} ~\cite{cordts2016cityscapes} focuses on capturing high variability of outdoor street scenes in common weather conditions from different cities. It contains 2,975 training images and 500 validation images with dense pixel-level labels. We transform the instance segmentation annotations into bounding boxes for our experiments.
	\item \textbf{Foggy Cityscapes}~\cite{sakaridis2018semantic} is built upon the images in the Cityscapes dataset~\cite{cordts2016cityscapes}. This dataset simulates the foggy weather using depth maps provided in Cityscapes with three levels of foggy weather, and thus is suitable to conduct weather adaptation experiments.
	\item \textbf{BDD100k}~\cite{yu2018bdd100k} consists of 100k images, with 70k training images and 10k validation images annotated with bounding boxes. %This dataset is associated with 6 types of weather, 6 different scenes, 3 categories for the time of day and 10 object categories.% 
	We extract a subset of BDD100k with images labeled as \emph{daytime}, including 36,728 training and 5,258 validation images. We use this subset for scene adaptation experiments.
	\item \textbf{PASCAL VOC}~\cite{everingham2010pascal} is a real-world dataset containing 20 categories of common objects with bounding box annotations. Following~\cite{saito2019strong-weak}, we employ PASCAL VOC 2007 and 2012 training and validation images (16,551 images in total) for experiments.
	\item \textbf{Clipart1k}~\cite{inoue2018cross-domain} contains 1k clipart images, which shares the same instance categories with PASCAL VOC but exhibits a large domain shift. We follow the practice in~\cite{saito2019strong-weak},  and use all images of Clipart1k for both training (without labels) and test.
\end{itemize}

\paragraph{Baselines and Comparison Methods.} We consider DA-Faster~\cite{chen2018domain} and the state-of-the-art SW-Faster~\cite{saito2019strong-weak} as our baseline methods, and re-implement them for fair comparisons. Our re-implementations achieve comparable or even better accuracies compared to the original papers. When comparing with other state-of-the-art methods, we report the results from original papers. Furthermore, we also train Faster R-CNN~\cite{ren2015faster} only using source images, as well as directly using annotated target images. We refer to models of these two settings as ``Faster R-CNN (Source)'' and ``Faster R-CNN (Oracle)'', respectively.

\paragraph{Implementation Details.} Following the default settings in~\cite{chen2018domain,saito2019strong-weak}, all training and test images are resized such that the shorter side has a length of $600$ pixels. By default, the backbone models are initialized using pre-trained weights of VGG-16~\cite{simonyan2015very} on ImageNet, but for the dissimilar domain adaptation experiments from PASCAL VOC~\cite{everingham2010pascal} to Clipart1k~\cite{inoue2018cross-domain}, we follow the practices in~\cite{saito2019strong-weak} and use ResNet-101~\cite{he2016deep} as the detection backbone. We fine-tune the network with a learning rate of $1 \times 10^{-3}$ for 50k iterations and then reduce the learning rate to $1 \times 10^{-4}$ for another 20k iterations. Each batch is composed of two images, one from source and another from target. The momentum of $0.9$ and the weight decay of $5 \times 10^{-4}$ is used for VGG-16 based detectors, while for ResNet-101 based detectors, we set the weight decay as $1 \times 10^{-4}$. In all experiments, we employ RoIAlign~\cite{he2017mask} for RoI feature extraction.

\subsection{Comparison Results}

\paragraph{Weather Adaptation.}
In real-world scenarios, object detectors may be applied under different weather conditions. We study the weather adaptation from clear weather to a foggy environment, using Cityscapes' training set and Foggy Cityscapes' validation set as the source domain and the target domain, respectively.

Table~\ref{tbl:weather_ad} shows the comparison results. Our categorical regularization framework can consistently boost the performance of DA-Faster and SW-Faster detectors, with 1.4\% and 2.6\% mAP improvements, respectively. In particular, our CCR module can greatly improve the detection results for some difficult categories such as ``train''. It clearly verifies the importance of increasing the weight of hard foreground instances in target domains for instance-level alignment. It is worth noting that our categorical regularization framework helps to reduce the performance gap between the domain adaptive detector and oracle detector trained with annotated target images to about 5\% mAP.

\begin{table*}[!ht]
\footnotesize
\centering
\caption{\textbf{Dissimilar Domain Adaptation}: Results on the Clipart1k dataset, using models trained on the PASCAL VOC training set.}
\begin{tabular}{l|p{0.2cm}<{\centering}p{0.2cm}<{\centering}p{0.2cm}<{\centering}p{0.2cm}<{\centering}p{0.25cm}<{\centering}p{0.2cm}<{\centering}p{0.2cm}<{\centering}p{0.2cm}<{\centering}p{0.2cm}<{\centering}p{0.2cm}<{\centering}p{0.2cm}<{\centering}p{0.2cm}<{\centering}p{0.2cm}<{\centering}p{0.33cm}<{\centering}p{0.35cm}<{\centering}p{0.22cm}<{\centering}p{0.25cm}<{\centering}p{0.2cm}<{\centering}p{0.2cm}<{\centering}p{0.3cm}<{\centering}|p{0.22cm}<{\centering}}
\toprule
Method & aero & bike & bird & boat & bottle & bus & car & cat & chair & cow & table & dog & horse & mbike & person & plant & sheep & sofa & train & tv & mAP \\
\hline
Faster R-CNN (Source)  &  21.9 & 42.2  & 22.9 & 19.0  & 30.8 & 43.1 & 28.9 &  10.7 & 27.4  & 18.1 & 13.5  & 10.3 & 25.0 & 50.7 &  39.0 & 37.4  & 6.9 & 18.1  & 39.2 & 34.9 & 27.0 \\
Kim~{\em{et~al.}}~\cite{kim2019self-training} &  28.0 & \textbf{64.5}  & 23.9 & 19.0  & 21.9 & 64.3 & \textbf{43.5} &  \textbf{16.4} & \textbf{42.2}  & 25.9 & \textbf{30.5}  & 7.9 & 25.5 & 67.6 &  54.5 & 36.4  & 10.3 & \textbf{31.2}  & \textbf{57.4} & 43.5 & 35.7 \\
\hline
DA-Faster~\cite{chen2018domain} & \textbf{38.0} & 47.5  & 27.7 & 24.8  & 41.3 & 41.2 & 38.2 &  11.4 & 36.8  & 39.7 & 19.6  & 12.7 & 31.9 & 47.8 &  55.6 & 46.3  & 12.1 & 25.6  & 51.1 & 45.5 & 34.7 \\
\textbf{DA-Faster-ICR (Ours)} & 31.0 & 53.9  & 29.2 & \textbf{28.2}  & \textbf{41.5} & 56.6 & 38.3 & 8.1 & 37.4  & 43.1 & 22.0 & 12.4 & 27.8 & 49.8 & 55.0 & \textbf{48.2}  & 11.0 & 22.7  & 54.2 & 46.9 & 35.9 \\
\textbf{DA-Faster-ICR-CCR (Ours)} & 30.2 & 57.0  & 30.6 & 26.2  & 38.0 & 57.1 & 36.1 &  12.7 & 36.4  & 44.8 & 18.2  & 14.6 & 30.0 & 56.7 &  56.6 & 45.9  & 17.8 & 25.3  & 50.5 & \textbf{48.5} & 36.7 \\
SW-Faster~\cite{saito2019strong-weak} & 29.2 & 53.1  & 30.2 & 24.4  & 41.4 & 52.5 & 34.6 &  14.0 & 36.3  & 43.5 & 17.6  & \textbf{16.6} & \textbf{33.4} & \textbf{78.1} &  59.1 & 42.1  & 15.8 & 24.9  & 45.5 & 43.7 & 36.8 \\
\textbf{SW-Faster-ICR (Ours)} & 25.2 & 54.0 & 31.7 & 23.4 & 40.3 & \textbf{65.8} & 35.4 & 12.1 & 37.6  & 48.1 & 18.6  & 14.2 & 31.3 & 73.6 & 59.9 & 46.5  & 19.5 & 25.9  & 46.0 & 45.6 & 37.7 \\
\textbf{SW-Faster-ICR-CCR (Ours)} & 28.7 & 55.3  & \textbf{31.8} & 26.0  & 40.1 & 63.6 & 36.6 &  9.4 & 38.7  & \textbf{49.3} & 17.6  & 14.1 & 33.3 & 74.3 &  \textbf{61.3} & 46.3  & \textbf{22.3} & 24.3  & 49.1 & 44.3 & \textbf{38.3} \\
\bottomrule
\end{tabular}
\label{tbl:dissimilar_ad}
\end{table*}

\begin{figure*}[!ht]
\centering
\includegraphics[width=1\textwidth]{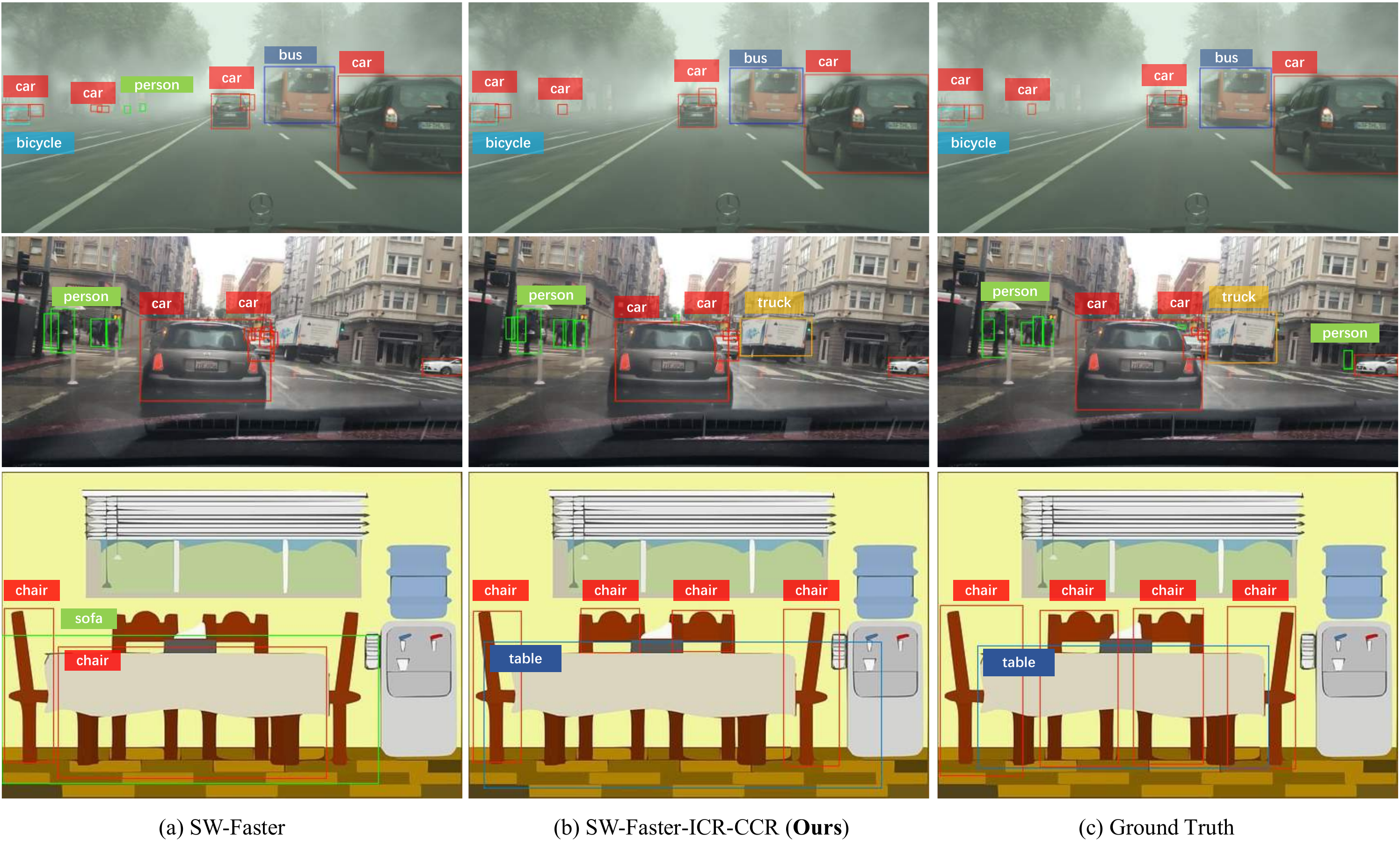}
\vspace{-2em}
\caption{Detection examples from three target datasets, from top to bottom: Foggy Cityscapes~\cite{sakaridis2018semantic}, BDD100k~\cite{yu2018bdd100k}, and Clipart1k~\cite{inoue2018cross-domain}. Our categorical regularization framework enables SW-Faster~\cite{saito2019strong-weak} to produce more accurate detection results with large domain shifts.}
\label{fig:det_examples}
\end{figure*}

\paragraph{Scene Adaptation.}

Scene layout changes frequently occur in real-life applications of object detection, \eg, automatic driving from one city to another. To study the effectiveness of our regularization framework for scene adaptation, we choose the Cityscapes~\cite{cordts2016cityscapes} training set as the source domain and a subset of BDD100k~\cite{yu2018bdd100k} as the target domain. In particular, we choose a subset of the BDD100k dataset annotated as \emph{daytime} to be our target domain and consider the city scene as the adaptation factor, since there only exists daytime data in the Cityscapes dataset. We report the detection results on seven common categories on both datasets.

As shown in Table~\ref{tbl:scene_ad}, we observe a significant performance gap between the domain adaptive detectors and the oracle detector, which suggests that scene layout shift is a challenging factor that hinders the performance of domain adaptive detection. Even under this difficult setting, our categorical regularization framework can also improve DA-Faster and SW-Faster by 1.3\% and 1.6\%, respectively. Similar to the observations on weather adaptation experiments, our CCR module can significantly improve the detection results of some difficult objects such as ``truck''.

\paragraph{Dissimilar Domain Adaptation.}

\begin{figure*}[!ht]
\centering
\includegraphics[width=1\textwidth]{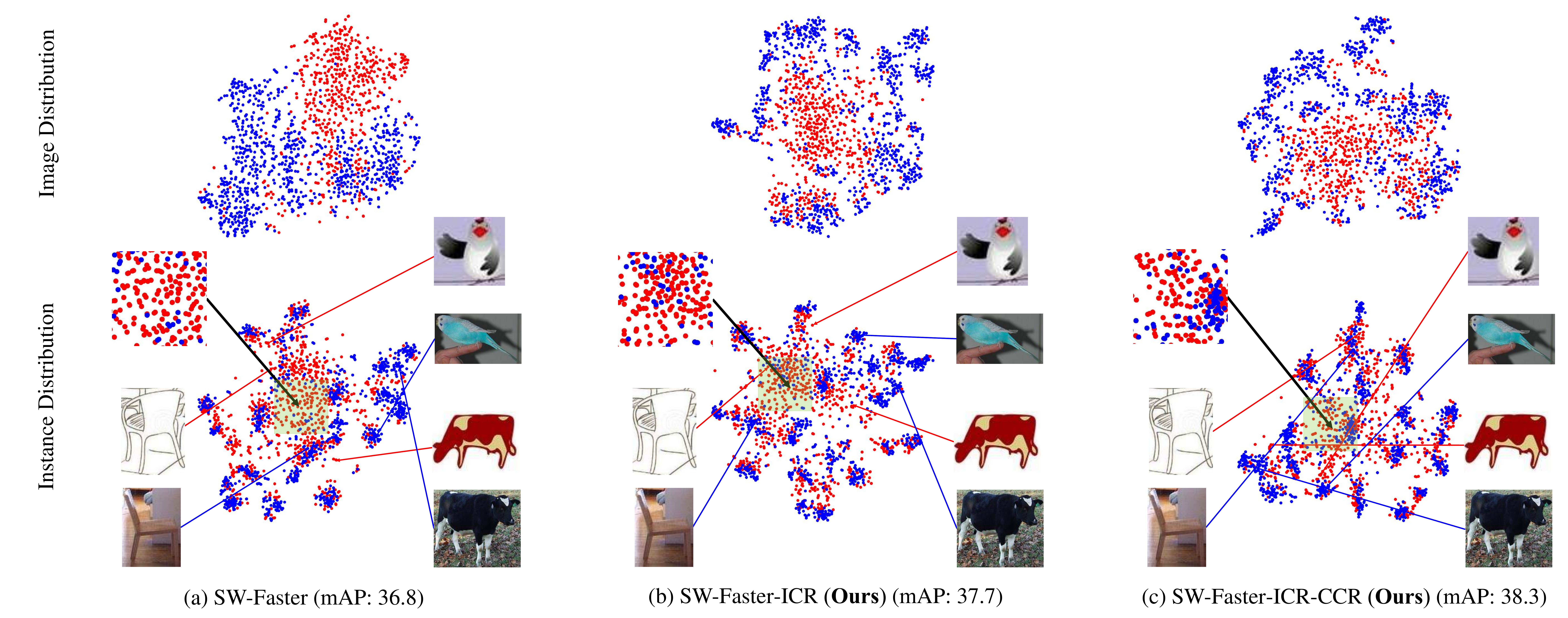}
\caption{Visualization of image features and instance features with $t$-SNE~\cite{maaten2008visualizing}, where the blue points represent source samples from PASCAL VOC~\cite{everingham2010pascal} and the red ones represent target samples from Clipart1k~\cite{inoue2018cross-domain}. \textbf{Top Row:} Holistic image features obtained by applying global average pooling to the output of the detection backbone network. \textbf{Second Row:} Instance features obtained by applying RoIAlign on the ground truth instances, where we also show three pairs of instances from different domains, and zoom in to the local regions of the most poorly matched instances. Compared to original SW-Faster~\cite{saito2019strong-weak}, our method better aligns both the image-level and instance-level features on both domains, and enables two dissimilar instances of the same category from different domains to stay close in the feature space.}
\label{fig:vis_distribution}
%\vspace{-1em}
\end{figure*}

Both weather adaptation and scene adaptation can be considered as adaptation between similar domains. We further show experiments on the dissimilar domain adaptation from real images to artistic images. We utilize Pascal VOC~\cite{everingham2010pascal} as the real source domain and the Clipart1k~\cite{inoue2018cross-domain} as the target domain. Clipart1k contains 1k comical images in total, which have the same 20 categories as PASCAL VOC. Following~\cite{saito2019strong-weak}, all images in Clipart1k are used for both training (without labels) and testing, and thus there is no oracle detector for this dataset.

As shown in Table~\ref{tbl:dissimilar_ad}, for dissimilar domain adaptation, our regularization framework also achieves considerable improvements over the baseline DA-Faster and SW-Faster by 2.0\% and 1.5\% mAP, respectively. Furthermore, our methods also outperform recent state-of-the-art one-stage object adaptive detector~\cite{kim2019self-training} that employs self training for domain adaptation.

\subsection{Visualization and Analyses}

\paragraph{Detection Examples.}
In Figure~\ref{fig:det_examples}, we show some detection examples from three target datasets, \ie, Foggy Cityscapes~\cite{sakaridis2018semantic}, BDD100k~\cite{yu2018bdd100k} and Clipart1k~\cite{inoue2018cross-domain}. Compared to the baseline SW-Faster~\cite{saito2019strong-weak} method, our SW-Faster-ICR-CCR method produces more accurate detection results under complex environments and large domain shifts.

\paragraph{Feature Visualization.}
We visualize the image and instance features learned for dissimilar domain adaptation (from PASCAL VOC~\cite{everingham2010pascal} to Clipart1k~\cite{inoue2018cross-domain}) using $t$-SNE~\cite{maaten2008visualizing}. For this experiment, we randomly sample 100 ground truth instances for each category, 50 from the source domain and 50 from the target domain. For some categories that have less than 50 instances in a certain domain, we sample all instances in that domain and the same number of instances from the other domain. The images containing these instances are sampled for image-level visualization. The image features are extracted by applying global average pooling on the output of the detection backbone network, while the instance features are extracted by RoIAlign.

As shown in Figure~\ref{fig:vis_distribution}, the blue points represent source samples and the red ones represent target samples. We also show three pairs of instances from different domains, and zoom in to the local regions of the most poorly matched instances. The dissimilar instance pairs of the same category from different domains stay closer in the feature space of our methods. Even for the most poorly matched region, our method still have better alignment performance than the baseline SW-Faster method~\cite{saito2019strong-weak}. Furthermore, thanks to the accurate instance-level alignment, our image-level alignment performance is also better than the baseline method.

\paragraph{Domain Distance.}
Besides visualization understanding, we also calculate a quantitative metric for domain distance, where both domains are represented by object instances. For this experiment, we use the same instance samples as the feature visualization experiment. Specifically, we adopt Earth Mover’s Distance (EMD)~\cite{rubner2000earth} as the metric for measuring domain distance. With this metric, domain distance computed for SW-Faster~\cite{saito2019strong-weak}, SW-Faster-ICR and SW-Faster-ICR-CCR are $8.84$, $8.59$, $8.15$, respectively.

The consistency between domain distance and model accuracy verifies the motivation of our work. That is, domain adaptive object detection relies heavily on aligning the crucial local regions and important instances on both domains. Our regularization framework assists the DA Faster R-CNN series to achieve this goal.

\section{Conclusions}

In this work, we presented a categorical regularization framework upon Domain Adaptive Faster R-CNN series for improving the adaptive detection performance. Specifically, we exploited the weakly localization ability of multi-label classification CNNs and the categorical consistency between image-level and instance-level predictions, which allows us to focus on aligning object-related local regions and hard aligned instances. In experiments, our framework significantly boosted the performance of existing Domain Adaptive Faster R-CNN detectors and produced state-of-the-art results on public benchmark datasets. Visualization and analyses can validate the effectiveness of our method. In the future, we will investigate how to apply our regularization framework to improve adaptive detectors beyond the Domain Adaptive Faster R-CNN series.

{\small
\bibliographystyle{ieee_fullname}
\bibliography{da_det}
}

\end{document}